%% file: main.tex
\documentclass[10pt,twocolumn,letterpaper]{article}

\usepackage[pagenumbers]{cvpr} 

\input{preamble}

\usepackage{stfloats}
\usepackage{pdflscape}
\definecolor{cvprblue}{rgb}{0.21,0.49,0.74}
\usepackage[pagebackref,breaklinks,colorlinks,citecolor=cvprblue]{hyperref}

\def\paperID{16310} 
\def\confName{CVPR}
\def\confYear{2024}

\title{A noisy elephant in the room: Is your out-of-distribution\\ detector robust to label noise?}

\author{Galadrielle Humblot-Renaux$^1$\qquad Sergio Escalera$^{1,2}$\qquad Thomas B. Moeslund$^1$\\%
$^1$Visual Analysis and Perception lab, Aalborg University, Denmark\\%
$^2$Department of Mathematics and Informatics \@ University of Barcelona and Computer Vision Center, Spain\\%
{\tt\small gegeh@create.aau.dk}\qquad {\tt\small sescalera@ub.edu}\qquad {\tt\small tbm@create.aau.dk}}

\begin{document}
\maketitle

\input{content_arxiv}

\end{document}

%% file: preamble.tex
\usepackage[dvipsnames,table,xcdraw]{xcolor}
\newcommand{\emin}[3]{$\epsilon_{\min}[$#1$>$#2$]=#3$}

\newcommand{\emininf}[3]{$\epsilon_{\min}[$#1$>$#2$]<#3$}
\newcommand{\emingeq}[3]{$\epsilon_{\min}[$#1$>$#2$]\geq#3$}
\newcommand{\eminexample}[2]{$\epsilon_{\min}[$#1$>$#2$]$}

\usepackage{adjustbox}
\usepackage{multirow}
\usepackage{booktabs}
\usepackage{makecell}
\usepackage{subcaption}
\usepackage{booktabs}
\usepackage{graphicx}
\usepackage[breakable,most]{tcolorbox}
\usepackage[normalem]{ulem}
\usepackage{xcolor}
\usepackage{soul}

\usepackage{amssymb}

\usepackage{array}
\newcommand{\PreserveBackslash}[1]{\let\temp=\\#1\let\\=\temp}
\newcolumntype{C}[1]{>{\PreserveBackslash\centering}p{#1}}
\newcolumntype{R}[1]{>{\PreserveBackslash\raggedleft}p{#1}}
\newcolumntype{L}[1]{>{\PreserveBackslash\raggedright}p{#1}}

\definecolor{lightgray}{gray}{0.9}

%% file: content_arxiv.tex
\begin{abstract}

The ability to detect unfamiliar or unexpected images is essential for safe deployment of computer vision systems. In the context of classification, the task of detecting images outside of a model's training domain is known as out-of-distribution (OOD) detection. While there has been a growing research interest in developing post-hoc OOD detection methods, there has been comparably little discussion around how these methods perform when the underlying classifier is not trained on a clean, carefully curated dataset. In this work, we take a closer look at 20 state-of-the-art OOD detection methods in the (more realistic) scenario where the labels used to train the underlying classifier are unreliable (e.g. crowd-sourced or web-scraped labels). Extensive experiments across different datasets, noise types \& levels, architectures and checkpointing strategies provide insights into the effect of class label noise on OOD detection, and show that poor separation between incorrectly classified ID samples vs. OOD samples is an overlooked yet important limitation of existing methods. Code: \small{\url{https://github.com/glhr/ood-labelnoise}}

\end{abstract}

\section{Introduction}
\label{sec:intro}

In many real-world applications where deep neural networks are deployed ``in the wild'', it is desirable to have models that not only correctly classify samples drawn from the distribution of labeled data but also flag unexpected inputs as out-of-distribution (OOD). This has motivated the development of a wide range of OOD detection methods and benchmarks for computer vision~\cite{unified-ood-survey-2022,generalized-ood-survey-2021}. In particular, \textit{post-hoc} OOD detection methods have shown wide appeal: compared to training-based methods, post-hoc OOD detectors can be applied on top of existing image classifiers without the need for re-training, have little to no architecture constraints, do not compromise classification performance, and achieve strong performance in large-scale settings~\cite{openood-benchmark-2022}.

Existing OOD benchmarks place significant emphasis on carefully designing the selection of OOD datasets used for evaluation~\cite{openood-benchmark-2022,imagenet-ood-benchmark-2023,cood-benchmark-2023,good-closed-set-is-all-you-need-2022}. In contrast, the role of the in-distribution (ID) dataset used for training the underlying classifier is seldom discussed. Among the most popular choices of ID dataset are MNIST, CIFAR10, CIFAR100 and ImageNet~\cite{openood-benchmark-2022,pytorch-ood-2022} - all of which have been carefully curated and reliably annotated. Yet, in practice, the collection of labelled datasets involves a trade-off between acquisition time/cost and annotation quality - human inattention, mis-clicking, limited expertise, crowd-sourcing, automated annotation, and other cost-saving measures inevitably introduce labelling errors~\cite{mturk-collection-2008}. Besides, some images are inherently ambiguous to label even for the most knowledgeable and careful of annotators~\cite{dl-noisy-labels-med-2020}. Considering how pervasive the problem of label noise is in real-world image classification datasets, its effect on OOD detection is crucial to study.

\begin{figure}[!t]
    \centering
    \includegraphics[height=3.2cm]{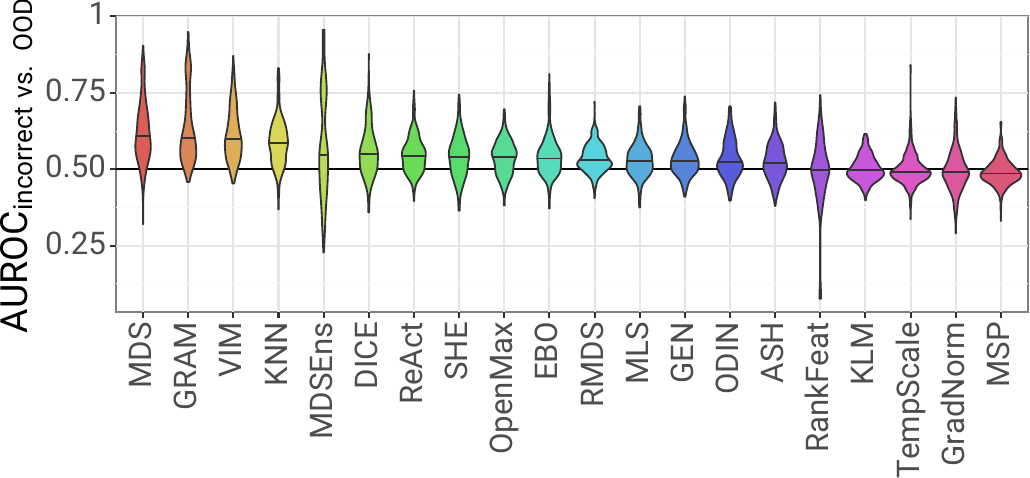}
    \caption{Can state-of-the-art OOD detectors tell incorrectly classified ID images apart from OOD inputs? Not really. Here we compare their performance across 396 trained classifiers.}
    \label{fig:intro-fig}
\end{figure}

To address this gap, we systematically analyse the label noise robustness of a wide range of OOD detectors, ranging from the widely adopted Maximum Softmax Probability (MSP) baseline~\cite{msp-2017,cood-benchmark-2023}, to distance-based methods operating in feature space~\cite{mds-2018,knn-2022}, to more recent, complex methods such as SHE~\cite{she-2023} and ASH~\cite{ash-2023}. In particular:
\begin{enumerate}
    \item We present the first study of post-hoc OOD detection in the presence of noisy classification labels, examining the performance of 20 state-of-the-art methods under different types and levels of label noise in the training data. Our study includes multiple classification architectures and datasets, ranging from the beloved CIFAR10 to the more difficult Clothing1M, and shows that even at a low noise rate, the label noise setting poses an interesting challenge for many methods.
    \item We revisit the notion that OOD detection performance correlates with ID accuracy~\cite{good-closed-set-is-all-you-need-2022,cood-benchmark-2023}, examining when and why this relation holds. Robustness to inaccurate classification requires that OOD detectors effectively separate mistakes on ID data from OOD samples - yet most existing methods confound the two (Figure~\ref{fig:intro-fig}).
    \item Our analysis includes key takeaways and recommendations for future evaluation and development of OOD detection methods considering an unreliable label setting.
\end{enumerate}

\section{Problem set-up}

In this work, we tackle the question: \textit{what happens when post-hoc OOD detectors are applied on top of a classifier trained with unreliable labels - a common setting in practice?} We introduce the main relevant concepts below.

\textbf{Classifier} We study OOD detection in the context of supervised image classification, where a discriminative model $h:\mathcal{X} \to \mathcal{Y} $ is trained on a dataset of $N$ labelled examples $D_{train} = \{(x_i,y_i)\}_{i=1}^N \in \mathcal{X} \times \mathcal{Y}$, where each $x$ is an input image and each $y$ is the corresponding class from the label space $\mathcal{Y}$. A common choice would for example be CIFAR10~\cite{cifar-dataset-2009}. $P_{train}(X,Y)$ defines the underlying training data distribution.  The classifier is evaluated on a test set $D_{test}$ drawn from the same distribution $P_{test}(X,Y) = P_{train}(X,Y)$. 

\textbf{OOD detector} Post-hoc OOD detection equips the trained classifier $h$ with a scoring function $o:X \to \mathbb{R} $ aiming to distinguish \textit{usual} examples drawn from $P_{test}(X)$ (ID samples) and \textit{anomalous} (OOD) examples drawn from a disjoint, held-out distribution $P_{out}(X)$. In practice, a collection of auxiliary datasets with minimal semantic overlap (e.g. CIFAR10 $\rightarrow$ SVHN~\cite{svhn-dataset-2011}) is commonly used for evaluation~\cite{openood-benchmark-2022}. Ideally, the score assigned to ID samples should be consistently lower (or higher) than for OOD samples, such that anomalous inputs can easily be flagged. 

\textbf{Label noise} We consider a noisy label setting, where the classifier $h$ does not have access to the true target values $y_i$ during training, but rather learns from a noisy dataset $D_{noisy} = \{(x_i,\hat{y}_i)\}_{i=1}^N$, where the target labels are corrupted: $\exists \, i$ such that $\hat{y}_i \neq y_i $. In this work, we consider only closed-set label noise, where $D_{noisy} \in \mathcal{X} \times \mathcal{Y}$ (that is, the noisy labels lie in the same label space as the true labels~\cite{evidentialmiax-openset-label-noise-2021}). The noise level $\epsilon$ is given by $P(y \neq \hat{y})$, the probability that an observed label is incorrect. Common models for studying and simulating label noise are (we refer to~\cite{classification-label-noise-survey-2013} for a detailed taxonomy): 
\begin{enumerate}
    \item Noisy Completely at Random (NCAR) or uniform label noise: labels are flipped at a constant rate $\epsilon$, regardless of class or image.
    \item Noisy at Random (NAR) or class-conditional label noise: a constant noise rate across all images of the same class, but different classes may have different noise rates.
    \item Noisy Not at Random (NNAR) or instance-dependent label noise: noisy labels are jointly determined by the true class and the associated image.
\end{enumerate}

In practice, \textit{real}  (as opposed to synthetically generated) label noise occurring from an imperfect annotation pipeline follows complex patterns, and is thus best represented by an instance-dependent model: some \textit{classes} are more likely to be mislabeled than others, and so are some \textit{images} (e.g. ambiguous or rare samples)~\cite{cifar-noisy-dataset-2022,animal10n-selfie-2019}.

\section{Related work}\label{sec:related}

\textbf{Studying label noise} The effect of unreliable labels on supervised learning is a well-studied problem in deep learning~\cite{deep-learning-noisy-labels-survey-2022} and computer vision~\cite{dl-noisy-labels-med-2020,image-classification-label-noise-survey-2021}, as errors or inconsistencies are a natural part of label collection in many real applications. Though increased dataset size can help~\cite{dl-robust-massive-labelnoise-2018}, noisy labels degrade classification performance, especially in the later stages of training where over-parameterized models are prone to memorizing them~\cite{rethinking-generalization-2017,earlylearning-regularization-memorization-noisylab-2020}. The precise effects of label noise have been shown to depend on the noise model and distribution~\cite{label-noise-distribution-robustness-2022}. A recent CIFAR classification benchmark suggests that models trained on real, instance-dependent noisy labels are significantly more prone to memorization than those trained on synthetic class-conditional labels with the same overall noise rate~\cite{cifar-noisy-dataset-2022}. We therefore consider real noisy labels in our benchmark (stemming from human annotation error), which we compare to two sets of synthetic noisy labels (uniform and class-conditional). We also compare the effect of validation-based early stopping vs. converging on the training set.

\textbf{Effect of label noise on reliability} Existing studies of label noise are largely focused on classification accuracy, and few works address the other side of the coin: \textit{reliability}. We look at reliability from the angle of OOD detection performance - to the best of our knowledge, there is currently no comparable study of OOD detection under a noisy label setting. Most closely related to our work are perhaps the experiments in~\cite{noisy-labels-dropout-de-2022} and the analysis in~\cite{noisy-labels-reliability-2022}. ~\cite{noisy-labels-dropout-de-2022} evaluates the effect of synthetic uniform label noise on MC-dropout and deep ensembles' uncertainty estimates, showing a significant degradation in OOD detection performance with increasing noise levels - in comparison, we study post-hoc OOD detection (with a wider variety of architectures, datasets, and methods) and consider real noisy datasets. ~\cite{noisy-labels-reliability-2022} studies label noise robustness in terms of model calibration, showing that early stopping, while beneficial in terms of accuracy, offers no reliability guarantees.

\textbf{Benchmarking OOD detection robustness} Previous works have investigated the limits of state-of-the-art OOD detection methods in various challenging settings, such as semantic similarity between ID vs. OOD classes~\citep{ood-limits-2021,cood-benchmark-2023,good-closed-set-is-all-you-need-2022}, fine-grained ID labels~\citep{mos-ood-2021}, large-scale datasets~\citep{scaling-ood-realworld-2022} and adversarial attacks~\cite{adverserial-attacks-2019,open-world-robustness-adv-2019}. In contrast, we focus on robustness to \textit{degraded classification performance on the ID dataset} due to noisy labels, which to the best of our knowledge has comparably received little attention.

\textbf{Relation between ID classification and OOD detection performance} In the standard clean label setting, a strong relationship between ID classification and OOD detection performance has been observed in prior work. ~\cite{good-closed-set-is-all-you-need-2022} studies the relation between closed-set (ID) classification and open-set recognition performance (AUROC), and finds open-set recognition performance to be highly correlated with classification accuracy. \cite{cood-benchmark-2023} observes a similar trend for out-of-distribution detection performance across a large variety of pre-trained deep learning architectures, using the MSP as OOD score. Both works only consider clean training datasets, and a small subset of methods. We study the extent to which this relation holds across a wider range of OOD detection methods and noisy datasets, and provide a very simple explanation for why some methods like MSP reach such a high correlation.

\section{OOD detection methods}

We evaluate 20 post-hoc OOD detection methods from the OpenOOD benchmark~\cite{openood-benchmark-2022} - currently the most comprehensive open-source benchmark available. Here we present and broadly categorize these methods based on how their scoring function is designed. 

\textbf{Softmax-based OOD detection} revolves around the idea that ID samples are associated with higher-confidence, lower-entropy predictions than OOD samples. The baseline Maximum Softmax Probability (\underline{MSP})~\cite{msp-2017} simply takes the Softmax ``confidence'' of the predicted class as OOD score. While MSP implicitly assumes a Softmax temperature of 1, \underline{TempScaling}~\cite{tempscaling-2017} treats the temperature as a hyper-parameter, softening or sharpening the Softmax probabilities (essentially modulating categorical entropy), with the aim of improving calibration. \underline{ODIN}~\cite{odin-2018} combines temperature scaling with input perturbation - ``pushin'' the input image a little in the direction that increases the MSP. In contrast, the Generalized ENtropy (\underline{GEN}) score considers the full predictive distribution and captures how much it deviates from a one-hot distribution.

\textbf{Logit-based OOD detection} bypasses the squashing effect of Softmax normalization. The Maximum Logit Score (\underline{MLS})~\cite{mls-2022} directly takes the logit of the predicted class. In a similar vein, energy-based OOD detection (\underline{EBO}) was first proposed in~\cite{ebo-2020}: a score is derived by applying the LogSumExp function to the logits - essentially a smooth version of the MLS, with an additional temperature parameter. Several post-hoc methods using an energy score have since followed suit, proposing various modifications to the network~\cite{dice-2022} or features~\cite{react-2021,rankfeat-2022,ash-2023} before extracting an energy score: \underline{REACT}~\cite{react-2021} clips activations at an upper bound, \underline{RankFeat}~\cite{rankfeat-2022} subtracts the rank-1 matrix from activations, \underline{DICE}~\cite{dice-2022} applies weight sparsification such that only the strongest contributors remain, and \underline{ASH}~\cite{ash-2023} sparsifies activations based on a pruning percentile.

\textbf{Distance-based OOD detection} aims to capture how much a test sample deviates from the ID dataset. The Mahalanobis distance score (\underline{MDS})~\cite{mds-2018} method fits a Gaussian distribution to the features of each class in the ID dataset; at test-time, the OOD score is taken as the distance to the closest class. The same authors also proposed \underline{MDSEnsemble}~\cite{mds-2018}, which computes an MDS score not just from the features extracted before the final layer, but also at earlier points in the network, and aggregates them. Alternatively, the Relative Mahalanobis distance score (\underline{RMDS})~\cite{rmds-2021} was proposed as a simple fix to MDS, which additionally fits a class-independent Gaussian to the entire ID dataset to compute a background score which is subtracted from the class-specific MDS score. Among other distance-based methods which rely on class-wise statistics, KLMatching (\underline{KLM})~\cite{mls-2022} takes the smallest KL Divergence between a test sample's Softmax probabilities and the mean Softmax probability vector for each ID class. \underline{OpenMax}~\cite{openmax-2016} operates in logit space, fitting a class-wise Weibull distribution to the distances of ID samples from the mean logit vector. Simplified Hopfield Energy (\underline{SHE})~\cite{she-2023} computes the inner product between a test sample's features and the mean ID feature of the predicted class. \underline{GRAM}~\cite{gram-2020} computes the Gram matrices of intermediate feature representations throughout the network, comparing them with the range of values observed for each class in the ID data. In contrast, deep k-nearest neighbor (\underline{KNN})~\cite{knn-2022} proposes a simple approach with no distributional assumptions - computing its score as the Euclidean distance to the closest samples from the ID set, regardless of class. Lastly, Virtual-logit Matching (\underline{VIM})~\cite{vim-2022} combines a logit energy score with a class-agnostic term capturing how features deviate from a principal subspace defined by the training set.

\textbf{Gradient-based OOD detection}: \underline{GradNorm}~\cite{gradnorm-2021} is the only method in OpenOOD which directly derives its score from the gradient space, claiming that gradient magnitude is higher for ID inputs.  The KL divergence between predicted Softmax probabilities and a uniform target is backpropagated to obtain gradients w.r.t the last layer parameters, followed by an $L_1$ norm to obtain the magnitude.

\section{Experiments}\label{sec:exp-setup}

We summarize our experimental set-up below, and refer to the supplementary for further details.

\textbf{ID Datasets} We select popular image classification datasets from the label noise literature featuring real noisy labels alongside clean reference labels, spanning different input sizes, number of classes, and sources \& levels of label noise - see Table~\ref{tab:dataset-overview}. The recently released CIFAR-N dataset~\citep{cifar-noisy-dataset-2022} provides noisy re-annotations of CIFAR-10 and CIFAR-100 collected via crowd-sourcing: each image was annotated by 3 people, and different noisy label sets were created for different label selection methods (majority voting, random selection, or worst label selection). Note that CIFAR-100-Fine and CIFAR-100-Coarse contain the same set of images - only the class definitions and labels differ. Clothing1M~\cite{clothing1m-dataset-2015} is a large-scale dataset collected by scraping shopping websites. Although the raw Clothing1M contains over a million images, we consider the smaller subset of images for which there is both a noisy and clean label.

\begin{table}[t!]
    \centering
    \begin{adjustbox}{max width=1\columnwidth}
    \begin{tabular}{rrcccc}
    \toprule
        & \textbf{\small{ID dataset}} & \textbf{\small{classes}} & \textbf{\small{\# images (train/val/test)}} & \textbf{\small{resolution}} & \textbf{\small{noise rate}}  \\ \midrule
        ~\cite{cifar-dataset-2009} & \textbf{CIFAR-10} & \multirow{4}{*}{10} &  \multirow{4}{*}{50,000/1,000/9,000} & \multirow{4}{*}{32x32} & 0\% \\
        \multirow{3}{*}{\cite{cifar-noisy-dataset-2022}} & CIFAR-10N-Agg & &  & & 9.01\% \\
        & CIFAR-10N-Rand1 &  & &   & 17.23\% \\
        & CIFAR-10N-Worst &  & &   & 40.21\%  \\ \midrule
        \multirow{1}{*}{\cite{cifar-dataset-2009}} & \textbf{CIFAR-100-Fine} &  \multirow{2}{*}{100} & \multirow{2}{*}{50,000/1,000/9,000} & \multirow{2}{*}{32x32} &  0\% \\ 
         \multirow{1}{*}{\cite{cifar-noisy-dataset-2022}} & CIFAR-100N-Fine &  & & & 40.20\% \\ \midrule
        \multirow{1}{*}{\cite{cifar-dataset-2009}}  & \textbf{CIFAR-100-Coarse} &  \multirow{2}{*}{20} & \multirow{2}{*}{50,000/1,000/9,000} & \multirow{2}{*}{32x32} & 0\% \\ 
        
        \multirow{1}{*}{\cite{cifar-noisy-dataset-2022}} & CIFAR-100N-Coarse &  & &   & 26.40\%  \\ \midrule
        \multirow{2}{*}{\cite{clothing1m-dataset-2015}} & \textbf{Clothing1M} & \multirow{2}{*}{14} & \multirow{2}{*}{24,637/7,465/5,395} & \multirow{2}{*}{256x256} & 0\% \\
        & Clothing1M-Noisy &  & &   & 38.26\%  \\ \bottomrule
    \end{tabular}
    \end{adjustbox}
    \caption{Dataset overview. Clean ones are shown in bold. The training set (clean or noisy labels) is used to train the classifier; the validation set (clean labels) is used for early stopping; the test set (clean labels) is used for evaluating classification and OOD detection performance. We always use clean labels for evaluation.}
    \label{tab:dataset-overview}
\end{table}

\textbf{Synthetic noise} For each real noisy label set, using the corresponding clean labels, we additionally create 2 synthetic counterparts with the same overall noise rate: one following a uniform (NCAR, class-independent) label noise model, and the other following a class-conditional label noise model with the exact same noise transition matrix as the real noise. We name these synthetic variants SU (Synthetic Uniform noise) and SCC (Synthetic Class-Conditional noise) - for example, from CIFAR-10N-Agg we create 2 synthetic versions, SU and SCC.

\textbf{OOD Datasets} For fair comparison, we use the same selection of OOD datasets for all models - the OOD datasets are therefore chosen such that there is minimal semantic overlap with any of the ID datasets. We include MNIST~\cite{mnist-dataset-2012}, SVHN~\cite{svhn-dataset-2011}, Textures~\cite{textures-2014} as they are commonly used as examples of ``far''-OOD~\cite{openood-benchmark-2022} (very different appearance and semantics than the ID dataset). As examples of more natural images, we also include EuroSAT-RGB~\cite{eurosat-2019}, Food-101~\cite{food101-2014}, a sub-set of the Stanford Online Products~\cite{stanford-online-products-2016}, and a 12-class sub-set of ImageNet. Since some methods require an OOD validation set for hyperparameter tuning, half of these classes are randomly selected and held-out for this purpose. The other 6 ImageNet classes, and the other OOD datasets make up the OOD test set.

\textbf{Evaluation metrics} We evaluate OOD detectors' ability to separate ID vs. OOD samples in terms of the Area Under the Receiver Operating Characteristic Curve (AUROC), where images from the ID test set (e.g. CIFAR10 test set) are considered positive samples, and those from the OOD test set (e.g. SVHN test set) as negatives. This is the most commonly reported metric in the literature~\cite{beyond-auroc-2023}, and we denote it as AUROC\textsubscript{ID vs. OOD}. In addition, unlike previous works, we separately measure the AUROC\textsubscript{correct vs. OOD} (and AUROC\textsubscript{incorrect vs. OOD}), where only correctly (or incorrectly) classified samples from the ID test set are considered - ideally, performance should be high on both metrics.

\textbf{Architectures} We include 3 architecture families: CNNs, MLPs and transformers.  We select lightweight architectures which have shown competitive results when trained on small-scale datasets: ResNet18~\cite{resnet-2016}, MLPMixer~\cite{mlp-mixer-2021} and Compact Transformers~\cite{compact-transformers-2021}. Following the OpenOOD benchmark~\cite{openood-benchmark-2022}, we do not adopt any advanced training strategies besides standard data augmentation. For each training dataset, we repeat training with 3 random seeds, and save 2 model checkpoints: an \textit{early} checkpoint (based on best validation accuracy) and the \textit{last} checkpoint (after a pre-defined number of epochs has elapsed, allowing for convergence - differs per architecture).

\textbf{Bird's eye view} To summarize, we train 3 different classifier architectures on 22 datasets (4 clean, 6 with real label noise, 12 with synthetic label noise), with 3 random seeds and 2 checkpoints saved per model - adding up to 396 distinct classifiers. On top of each classifier, 20 different OOD detection methods are applied and evaluated on 7 OOD datasets.  Throughout the paper, OOD detection performance is taken as the median across the 7 OOD datasets (see the supplementary for results and a discussion of the median vs. mean OOD detection performance).

\textbf{Statistical significance tests} When comparing pairs of  methods or settings, we use the Almost Stochastic Order (ASO) test \citep{del2018optimal, dror2019deep} as implemented by \citet{ulmer2022deep}. This statistical test was specifically designed to compare deep learning models, making no distributional assumptions. We apply ASO with a significance level $\alpha = 0.05$ and report \eminexample{A}{B}. If \emingeq{A}{B}{0.5} we cannot claim that method A is better than method B; the smaller $\epsilon_{\min}$, the more confident we can be that method A is superior.

\begin{table*}[!t]
\centering
\resizebox{0.92\textwidth}{!}{%
\begin{tabular}{l|c|ccc|ccc|ccc|cccc|cccc|cccc}
\multicolumn{1}{c|}{\multirow{2}{*}{\textit{\begin{tabular}[c]{@{}c@{}}training\\ labels\end{tabular}}}} & \multicolumn{10}{c|}{CIFAR10} & \multicolumn{4}{c}{\multirow{2}{*}{CIFAR100-Coarse}} & \multicolumn{4}{c|}{\multirow{2}{*}{CIFAR100-Fine}} & \multicolumn{4}{c}{\multirow{2}{*}{Clothing1M}} \\
\multicolumn{1}{c|}{}  & \multicolumn{1}{l}{} & \multicolumn{3}{c}{Agg} & \multicolumn{3}{c}{Rand1} & \multicolumn{3}{c|}{Worst} & \multicolumn{4}{c}{} & \multicolumn{4}{c|}{} & \multicolumn{4}{c}{} \\ \hline
\multicolumn{1}{c|}{\textit{method}} &  \textbf{clean} & \textbf{N} & \textbf{SCC} & \textbf{SU} & \textbf{N} & \textbf{SCC} & \textbf{SU} & \textbf{N} & \textbf{SCC} & \textbf{SU} & \textbf{clean} & \textbf{N} & \textbf{SCC} & \textbf{SU} & \textbf{clean} & \textbf{N} & \textbf{SCC} & \textbf{SU} & \textbf{clean} & \textbf{N} & \textbf{SCC} & \textbf{SU} \\ \hline
GRAM & \textbf{94.45} & \textbf{89.49} & \textbf{89.12} & \textbf{90.7} & 88.82 & \textbf{89.19} & {\ul \textbf{90.52}} & {\ul \textbf{88.6}} & {\ul \textbf{88.73}} & \textbf{87.98} & 82.07 & {\ul \textbf{80.05}} & \textbf{82.1} & {\color[HTML]{ED7D31} \textbf{79.2}} & 82.93 & {\color[HTML]{ED7D31} \textbf{76.31}} & {\ul \textbf{80.24}} & {\ul \textbf{82.64}} & \textbf{91.04} & \textbf{89.07} & \textbf{94.71} & \textbf{95.37} \\
MDS & {\ul \textbf{96.07}} & 87.93 & {\ul \textbf{92.4}} & {\ul \textbf{92.97}} & {\ul \textbf{92.37}} & {\ul \textbf{89.25}} & \textbf{87.45} & \textbf{86.74} & \textbf{86.49} & {\ul \textbf{89.2}} & 80.07 & {\color[HTML]{ED7D31} \textbf{78.89}} & {\ul \textbf{82.84}} & {\ul \textbf{80.12}} & 80.1 & {\color[HTML]{9C0006} 74.96} & {\color[HTML]{9C0006} 74.48} & {\color[HTML]{9C0006} 73.49} & 87.12 & \textbf{90.98} & \textbf{88.58} & \textbf{92.38} \\
VIM & \textbf{95.65} & \textbf{89.9} & \textbf{91.81} & \textbf{92.3} & 88.75 & \textbf{88.6} & 84.49 & \textbf{86.31} & \textbf{87.23} & \textbf{88.75} & {\ul \textbf{84.29}} & {\color[HTML]{ED7D31} 76.61} & 80.04 & {\color[HTML]{ED7D31} \textbf{78}} & 81.37 & {\color[HTML]{ED7D31} 75.31} & {\color[HTML]{9C0006} 73.24} & {\color[HTML]{9C0006} \textbf{73.94}} & \textbf{88.99} & 83.09 & 87.17 & 90.14 \\
MDSEns & 92.57 & 83.89 & 83.62 & 83.79 & 81.8 & 83.06 & 80.36 & 82.95 & 84.02 & 84.11 & {\color[HTML]{ED7D31} 79.25} & {\color[HTML]{ED7D31} \textbf{78.31}} & {\color[HTML]{ED7D31} 77.41} & {\color[HTML]{9C0006} 73.6} & {\ul \textbf{84.85}} & {\color[HTML]{ED7D31} {\ul \textbf{77.47}}} & {\color[HTML]{ED7D31} \textbf{78.43}} & {\color[HTML]{ED7D31} \textbf{79.85}} & {\ul \textbf{95.36}} & {\ul \textbf{95.44}} & {\ul \textbf{95.78}} & {\ul \textbf{95.69}} \\
KNN & 93.63 & {\ul \textbf{90.07}} & 88.75 & 90.14 & 87.74 & 86.66 & 85.11 & 86.3 & 83.73 & 84.14 & \textbf{84.16} & {\color[HTML]{9C0006} 74.4} & \textbf{80.39} & {\color[HTML]{ED7D31} 75.82} & 83.29 & {\color[HTML]{ED7D31} 75.67} & {\color[HTML]{ED7D31} \textbf{76.48}} & {\color[HTML]{9C0006} 71.35} & 85.32 & 85.5 & 84.59 & 80.87 \\
RMDS & 92.92 & 89.38 & 87.94 & 88.14 & 89.07 & 85.73 & \textbf{87.04} & 84.03 & 81.99 & 82.35 & 82.14 & {\color[HTML]{ED7D31} 75.93} & {\color[HTML]{ED7D31} 77.36} & {\color[HTML]{9C0006} 74.81} & 83.28 & {\color[HTML]{ED7D31} 76} & {\color[HTML]{ED7D31} 75.75} & {\color[HTML]{9C0006} 73.48} & {\color[HTML]{ED7D31} 75.81} & {\color[HTML]{9C0006} 71.43} & {\color[HTML]{ED7D31} 78.22} & {\color[HTML]{9C0006} 66.66} \\
DICE & 90 & 83.33 & 84.18 & 86.24 & 88.52 & 81 & 86.21 & 82.79 & {\color[HTML]{ED7D31} 79.58} & {\color[HTML]{ED7D31} 79.07} & \textbf{82.79} & {\color[HTML]{ED7D31} 77.68} & {\color[HTML]{ED7D31} 75.01} & {\color[HTML]{9C0006} 70.43} & 82.52 & {\color[HTML]{ED7D31} \textbf{76.51}} & {\color[HTML]{9C0006} 73.92} & {\color[HTML]{9C0006} 68.41} & 84.96 & {\color[HTML]{ED7D31} 75.72} & 86.69 & 82.89 \\
ReAct & 90.91 & 87.32 & 86.63 & 82.16 & 89.74 & 82.96 & 81.97 & 84.5 & {\color[HTML]{ED7D31} 78.41} & 80.11 & \textbf{82.79} & {\color[HTML]{9C0006} 73.09} & {\color[HTML]{9C0006} 73.62} & {\color[HTML]{9C0006} 70.38} & \textbf{83.76} & {\color[HTML]{9C0006} 73.57} & {\color[HTML]{9C0006} 73.71} & {\color[HTML]{9C0006} 67.55} & 82.57 & {\color[HTML]{9C0006} 73.1} & 80.22 & {\color[HTML]{ED7D31} 76.58} \\
GEN & 91.86 & 85.99 & 85.44 & 82.08 & \textbf{89.86} & 82.89 & 80.84 & 83.75 & 81.81 & 80.57 & 82.69 & {\color[HTML]{9C0006} 73.25} & {\color[HTML]{9C0006} 71.47} & {\color[HTML]{9C0006} 70.99} & 81.34 & {\color[HTML]{9C0006} 73.4} & {\color[HTML]{9C0006} 73.1} & {\color[HTML]{9C0006} 67.11} & 83.91 & {\color[HTML]{9C0006} 73.57} & {\color[HTML]{ED7D31} 79.79} & {\color[HTML]{ED7D31} 76.78} \\
EBO & 91.31 & 84.87 & 85.73 & 81.62 & \textbf{89.88} & 81.93 & {\color[HTML]{ED7D31} 77.88} & 83.04 & 81.38 & {\color[HTML]{ED7D31} 77.4} & 82.74 & {\color[HTML]{9C0006} 72.99} & {\color[HTML]{9C0006} 70.93} & {\color[HTML]{9C0006} 67.85} & 81.41 & {\color[HTML]{9C0006} 73.65} & {\color[HTML]{9C0006} 73.01} & {\color[HTML]{9C0006} 67.42} & 85.19 & {\color[HTML]{ED7D31} 76.32} & 85.31 & {\color[HTML]{ED7D31} 76.64} \\
SHE & 89.6 & 87.81 & 84.33 & 86.48 & 86.63 & 83.16 & 83.04 & 83.24 & 80.06 & {\color[HTML]{ED7D31} 78.98} & 80.42 & {\color[HTML]{9C0006} 71.8} & {\color[HTML]{ED7D31} 80} & {\color[HTML]{9C0006} 70.11} & 80.38 & {\color[HTML]{9C0006} 69.63} & {\color[HTML]{9C0006} 69.56} & {\color[HTML]{9C0006} 66.68} & 82.29 & {\color[HTML]{ED7D31} 78.07} & {\color[HTML]{ED7D31} 78.4} & {\color[HTML]{ED7D31} 77.73} \\
ODIN & 91.47 & 87.71 & 86.31 & 82.48 & 89.79 & 82.21 & 80.68 & 84.09 & 82.13 & 80.09 & 81.42 & {\color[HTML]{9C0006} 73.1} & {\color[HTML]{9C0006} 70.88} & {\color[HTML]{9C0006} 69.97} & \textbf{83.59} & {\color[HTML]{9C0006} 74.85} & {\color[HTML]{9C0006} 72.19} & {\color[HTML]{9C0006} 67.16} & 83.38 & {\color[HTML]{9C0006} 71.59} & {\color[HTML]{ED7D31} 77.73} & {\color[HTML]{ED7D31} 75.47} \\
MLS & 91.26 & 84.76 & 85.59 & 81.57 & 88.81 & 82.31 & {\color[HTML]{ED7D31} 78} & 83.54 & 82.01 & 80.01 & 82.66 & {\color[HTML]{9C0006} 72.92} & {\color[HTML]{9C0006} 70.85} & {\color[HTML]{9C0006} 69.19} & 81.45 & {\color[HTML]{9C0006} 73.64} & {\color[HTML]{9C0006} 72.5} & {\color[HTML]{9C0006} 67.03} & 83.3 & {\color[HTML]{9C0006} 72.75} & {\color[HTML]{ED7D31} 77.74} & {\color[HTML]{ED7D31} 75.54} \\
TempScale & 91.67 & 85.76 & 85.04 & 82.25 & 85.07 & {\color[HTML]{ED7D31} 78.1} & {\color[HTML]{ED7D31} 79.78} & 82.85 & 80.51 & 80.13 & 81.63 & {\color[HTML]{9C0006} 71.67} & {\color[HTML]{9C0006} 69.94} & {\color[HTML]{9C0006} 69.35} & 80.75 & {\color[HTML]{9C0006} 72.66} & {\color[HTML]{9C0006} 71.28} & {\color[HTML]{9C0006} 66.98} & {\color[HTML]{ED7D31} 79.82} & {\color[HTML]{9C0006} 68.77} & 86.28 & {\color[HTML]{9C0006} 74.45} \\
ASH & 88.33 & 84.75 & 82.37 & 81.66 & 82.29 & {\color[HTML]{ED7D31} 76.47} & {\color[HTML]{9C0006} 72.48} & 85.27 & {\color[HTML]{ED7D31} 78.26} & {\color[HTML]{ED7D31} 75.48} & 82.78 & {\color[HTML]{9C0006} 71.19} & {\color[HTML]{9C0006} 73.21} & {\color[HTML]{9C0006} 68.43} & 82.74 & {\color[HTML]{9C0006} 74.13} & {\color[HTML]{9C0006} 70.95} & {\color[HTML]{9C0006} 67.48} & 81.53 & {\color[HTML]{9C0006} 74.89} & {\color[HTML]{ED7D31} 76.63} & {\color[HTML]{ED7D31} 75.74} \\
OpenMax & 90.5 & 86.12 & 83.46 & 82.26 & 83.05 & 82.87 & {\color[HTML]{ED7D31} 79.12} & 80.39 & {\color[HTML]{ED7D31} 75.68} & {\color[HTML]{ED7D31} 77.41} & 81.14 & {\color[HTML]{ED7D31} 76.69} & {\color[HTML]{9C0006} 72.65} & {\color[HTML]{9C0006} 69.17} & 80.13 & {\color[HTML]{9C0006} 72.82} & {\color[HTML]{ED7D31} 75.6} & {\color[HTML]{9C0006} 67.66} & {\color[HTML]{9C0006} 71.74} & {\color[HTML]{9C0006} 69.23} & {\color[HTML]{9C0006} 72.55} & {\color[HTML]{9C0006} 74.36} \\
MSP & 91.34 & 85.19 & 84.87 & 82.41 & 85.13 & 82.21 & 80.68 & 82.48 & {\color[HTML]{ED7D31} 79.4} & 80.09 & 80.51 & {\color[HTML]{9C0006} 70.42} & {\color[HTML]{9C0006} 68.88} & {\color[HTML]{9C0006} 69.97} & {\color[HTML]{ED7D31} 79.85} & {\color[HTML]{9C0006} 70.92} & {\color[HTML]{9C0006} 69.61} & {\color[HTML]{9C0006} 66.92} & {\color[HTML]{ED7D31} 77.57} & {\color[HTML]{9C0006} 66.02} & {\color[HTML]{9C0006} 72.44} & {\color[HTML]{9C0006} 73.89} \\
KLM & 90.84 & 83.7 & 82.15 & 81.69 & 80.13 & 81.88 & 80.64 & {\color[HTML]{9C0006} 74.79} & {\color[HTML]{ED7D31} 75.93} & {\color[HTML]{ED7D31} 76.83} & {\color[HTML]{ED7D31} 79.37} & {\color[HTML]{9C0006} 70.2} & {\color[HTML]{9C0006} 69.24} & {\color[HTML]{9C0006} 67.81} & {\color[HTML]{ED7D31} 79.58} & {\color[HTML]{9C0006} 71.52} & {\color[HTML]{9C0006} 69.89} & {\color[HTML]{9C0006} 67.25} & {\color[HTML]{ED7D31} 77.26} & {\color[HTML]{9C0006} 65.49} & {\color[HTML]{9C0006} 65.22} & {\color[HTML]{9C0006} 62.01} \\
GradNorm & 86.22 & {\color[HTML]{ED7D31} 79.55} & {\color[HTML]{ED7D31} 77.1} & {\color[HTML]{ED7D31} 77.84} & 81.66 & {\color[HTML]{ED7D31} 79.88} & {\color[HTML]{ED7D31} 76.96} & 84.11 & {\color[HTML]{9C0006} 72.96} & {\color[HTML]{9C0006} 72.88} & {\color[HTML]{9C0006} 69.29} & {\color[HTML]{9C0006} 66.27} & {\color[HTML]{9C0006} 73.17} & {\color[HTML]{9C0006} 67.44} & {\color[HTML]{9C0006} 70.95} & {\color[HTML]{9C0006} 65.65} & {\color[HTML]{9C0006} 71.26} & {\color[HTML]{9C0006} 63.29} & {\color[HTML]{ED7D31} 79.71} & {\color[HTML]{ED7D31} 75.32} & {\color[HTML]{9C0006} 73.93} & {\color[HTML]{ED7D31} 77.52} \\
RankFeat & 81.83 & 83.53 & {\color[HTML]{ED7D31} 75.86} & {\color[HTML]{9C0006} 73.12} & {\color[HTML]{ED7D31} 78.29} & {\color[HTML]{ED7D31} 75.07} & {\color[HTML]{ED7D31} 79.14} & 82.57 & {\color[HTML]{ED7D31} 77.25} & {\color[HTML]{ED7D31} 75.75} & {\color[HTML]{9C0006} 73.15} & {\color[HTML]{9C0006} 64.6} & {\color[HTML]{9C0006} 69.54} & {\color[HTML]{9C0006} 65.44} & {\color[HTML]{9C0006} 68.77} & {\color[HTML]{ED7D31} 76.07} & {\color[HTML]{9C0006} 70.24} & {\color[HTML]{9C0006} 67.64} & {\color[HTML]{9C0006} 69.16} & {\color[HTML]{ED7D31} 75.85} & {\color[HTML]{9C0006} 69.93} & {\color[HTML]{9C0006} 73.37} \\
\hline
\end{tabular}%
}
\caption{\textbf{Best-case} OOD detection performance (AUROC\textsubscript{ID vs. OOD} in \%) per method (that is, after selecting the best architecture-seed-checkpoint combination for each training label set). N, SCC, and SU refer to the real and synthetic noisy label sets described in Section~\ref{sec:exp-setup}. The top-3 for each training dataset are highlighted in bold, and the top-1 is underlined. In red are scores $<75\%$ and in orange scores between 75 and 80\%. Rows are sorted based on the total performance across columns.}
\label{tab:best-ood-per-method}
\end{table*}

\section{Analysis}

We explore the effect of label noise on OOD detection, starting with an overall view of performance trends in Section~\ref{sec:overall-perf-effect}, then looking at OOD detection in relation to classification performance in Section~\ref{sec:accuracy-vs-ood}, delving into what works (and what doesn't) in Section~\ref{sec:what-works}, and raising important considerations about how/whether to make use of a clean validation set in Section~\ref{sec:val-stuff}. Section~\ref{sec:osr} extends results to a more practical setting.  More detailed analyses and additional supporting figures are in the supplementary.

\subsection{Where there's noise there's trouble} \label{sec:overall-perf-effect}

\begin{figure}[t]
    \centering
    \includegraphics[width=\columnwidth]{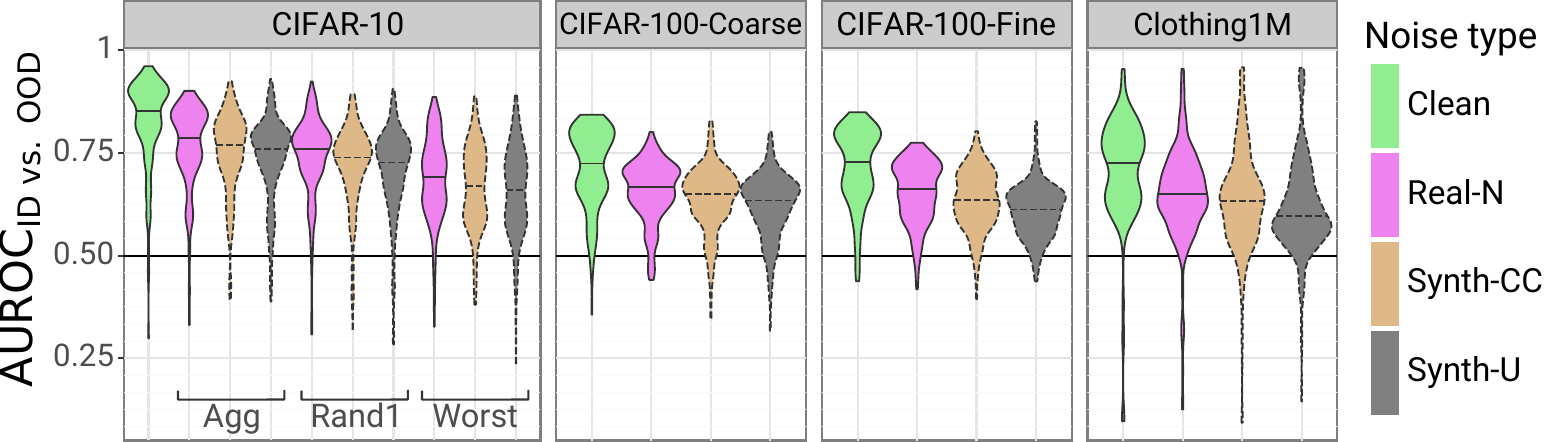}
    \caption{Distribution of OOD detection performance across methods \& models when training the classifier on different label sets.}
    \label{fig:auroc-gap-noisy-violins}
\end{figure}

Figure~\ref{fig:auroc-gap-noisy-violins} gives an overview of OOD detection performance for different training datasets and label noise settings. We see a clear drop in overall OOD detection performance when label noise is introduced in the training dataset, compared to training on a cleanly labelled dataset (green). Even with only 9\% of incorrect CIFAR10 labels  (CIFAR-Agg labels sets), the median $\text{AUROC}_{\text{ID vs. OOD}}$ across all models drops by over 5\%. In Table~\ref{tab:best-ood-per-method}, for each method we report the \textit{best-case} OOD detection performance for a given training label set. While most methods are able to reach 80\% $\text{AUROC}_{\text{ID vs. OOD}}$ with a classifier trained on clean labels, the number of competitive methods falls with increasing label noise, especially at noise rates $> 20\%$. GRAM, KNN, MDS, MDSEnsemble and VIM are the only methods able to reach 90+\% AUROC on at least one of the noisy datasets.

\noindent\ul{\textbf{Takeaway: Enter the elephant}}
Label noise in the classifier's training data makes it more difficult for post-hoc OOD detection methods to flag unfamiliar samples at test-time, even in small-scale settings like CIFAR10.

\subsection{Does accuracy tell the whole story?} \label{sec:accuracy-vs-ood}

The most obvious effect of label noise in the training data is a decrease in classification performance on ID test data. At the same time, previous works have remarked a strong relation between classification performance and OOD detection for popular post-hoc methods like MSP~\cite{cood-benchmark-2023} and MLS~\cite{good-closed-set-is-all-you-need-2022}. We dig deeper. When does this relation hold and why?

\begin{figure}[b]
    \centering
    \includegraphics[width=\columnwidth]{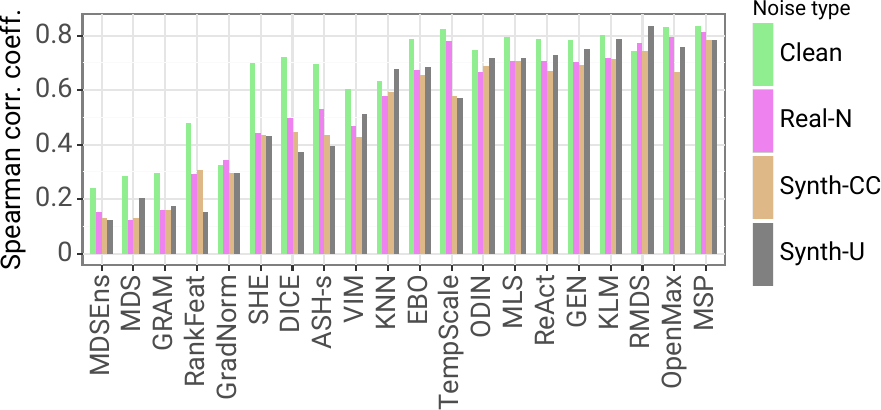}
    \caption{Does OOD detection performance (AUROC\textsubscript{ID vs. OOD}) correlate with ID classification performance (accuracy)? We measure the rank correlation across different architectures, seeds, checkpoints, and datasets for different label sets. All results shown here are statistically significant ($p << 0.001$).}
    \label{fig:accuracy_auroc_corr_per_noise_type}
\end{figure}

\textbf{For which methods does this relation hold?} In Figure~\ref{fig:accuracy_auroc_corr_per_noise_type},  we quantify the relationship between ID accuracy and AUROC\textsubscript{ID vs. OOD} in terms of Spearman correlation $\rho$. We find that correlation varies widely across methods, being the strongest for MSP, and is generally weaker for those which operate earlier in the network. We also note that for all methods except KNN and RMDS, the label noise setting makes OOD detection performance \textit{less predictable} - and so does early stopping (cf. Section~\ref{sec:val-stuff}). This points to the distribution of ID scores playing an important role in OOD detection performance.

\begin{figure}[t]
    \centering
    \includegraphics[width=\columnwidth]{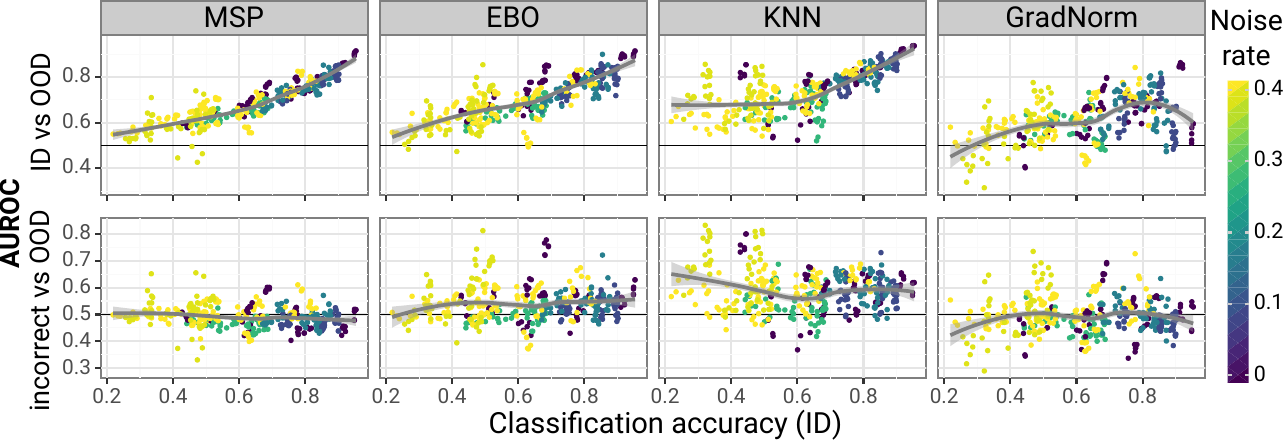}
    
    \caption{Relationship between ID classification performance and OOD detection performance, considering all ID test samples (top) or only incorrectly classified ones (bottom) in the AUROC metric. Each point corresponds to a single model.}
    
    \label{fig:accuracy-vs-ood-selected}
\end{figure}

\textbf{When it does - why?} We provide a simple observation which is lacking in prior work: methods whose OOD detection performance predictably degrades alongside classification accuracy are characterized by a high AUROC\textsubscript{correct vs. OOD} and a low AUROC\textsubscript{incorrect vs. OOD}. On clean, easy datasets like CIFAR10, they exhibit strong OOD detection performance as there are few incorrectly predicted ID samples in the test set (thus the AUROC\textsubscript{incorrect vs. OOD} term is negligible in the overall AUROC\textsubscript{ID vs. OOD}) - however, when the number of incorrect prediction grows, the low AUROC\textsubscript{incorrect vs. OOD} becomes a more significant factor. Importantly and as exemplified by Figure~\ref{fig:accuracy-vs-ood-selected}, for all methods, AUROC\textsubscript{incorrect vs. OOD} is not (or only weakly, $\rho < 0.2$) correlated with classification accuracy.  MSP is the most clear-cut example, with a median AUROC\textsubscript{incorrect vs. OOD} of around 0.5 across all dataset-architecture-seed-checkpoint combinations (bottom left of Figure~\ref{fig:accuracy-vs-ood-selected}) - that is, MSP often is no better (or worse) than a random detector at separating ID mistakes and OOD inputs, no matter how accurate the underlying classifier is. The Top-4 methods in Table~\ref{tab:best-ood-per-method} are the only ones with a median $\text{AUROC}_{\text{incorrect vs. OOD}} \geq 0.6$ - none of the other methods exceed a median $\text{AUROC}_{\text{incorrect vs. OOD}}$ of 0.55 - see Figure~\ref{fig:intro-fig}.

\noindent \ul{\textbf{Takeaway: Would your OOD detector be better off as a failure detector?}}
  Accuracy correlating with OOD detection performance is partly symptomatic of many seemingly effective methods being unable to separate \textit{incorrectly classified ID samples} from OOD samples - a bottleneck for robustness to imperfect classification. Claims that post-hoc OOD detection can be improved by simply improving the underlying classifier~\cite{good-closed-set-is-all-you-need-2022} overlook this fundamental issue.

\begin{figure}[t]
    \centering
    \includegraphics[width=\columnwidth]{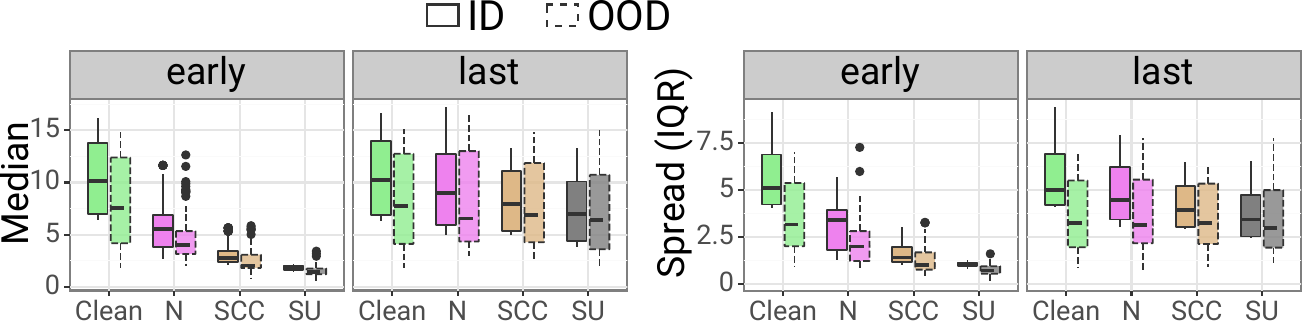}
    \caption{Max Logit ID and OOD score statistics across models trained on Clothing1M, for different noise types \& checkpointing.}
    \label{fig:noise-types-score-stats}
\end{figure}

\textbf{It's not just about the noise rate} We find that for a fixed noise rate in a given dataset, different types/models of label noise yield comparable classification accuracy ($\epsilon_{\min} \geq 0.5$ for all pair-wise comparisons), yet have different effects on OOD performance. Indeed, real label noise is better handled than the same level of synthetic by most methods, with SU labels being the most challenging - this trend is clear in Figure~\ref{fig:auroc-gap-noisy-violins}. Figure~\ref{fig:noise-types-score-stats} shows an example of how different noise types and checkpointing strategies shape the magnitude and spread of logits. Intuitively, when the noise is spread randomly across samples (SU noise model), it is more difficult to learn which kinds of images or classes to be uncertain about, leading to consistently lower-confidence predictions across all ID samples (low median, low spread). Conversely, when label noise is more concentrated for certain classes (SCC) and/or for certain features (real noise), the classifier can learn to be more confident in some parts of the input space than others (higher median, higher spread).

\noindent \uline{\textbf{Takeaway: Faking it is better than ignoring it}} Uniform (synthetic) label noise in the training data tends to degrade OOD detection more strongly than class-dependant (synthetic) and instance-dependent (real) label noise. We encourage the use of synthetic uniform labels to evaluate the worst-case performance of OOD detectors, as they can be easily generated for any image classification dataset. 
   
\begin{figure*}[!t]
    \centering
    \includegraphics[width=\textwidth]{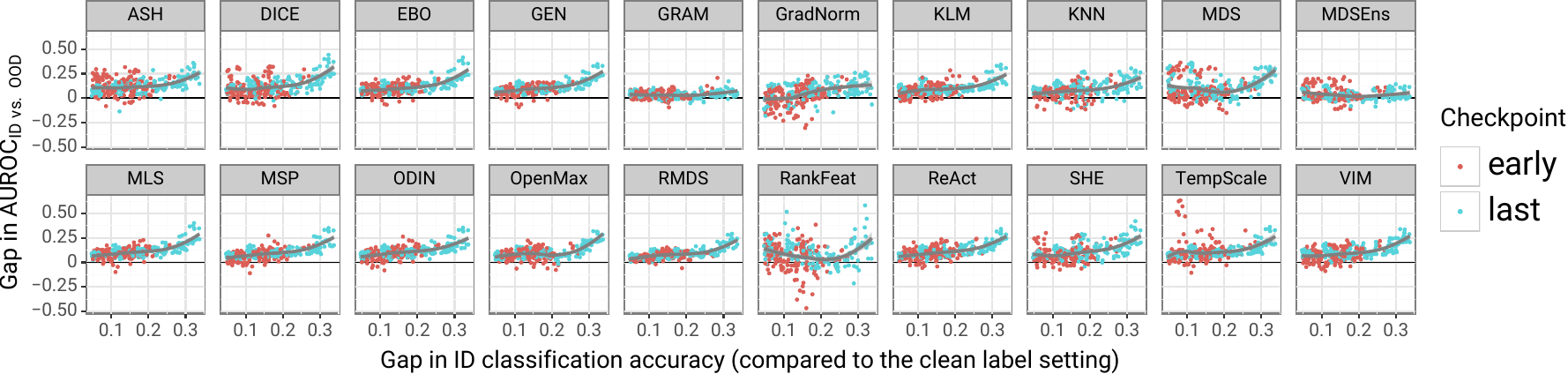}
    \caption{Relation between the drop in accuracy caused by noisy labels and the resulting drop in OOD detection performance across all 20 methods. Each point corresponds to a single model trained with noisy labels. }
    \label{fig:auroc_gap-vs-accuracy_gap_per_checkpoint}
\end{figure*}

\subsection{Design features which hurt or help} \label{sec:what-works}

\textbf{Why are the winners the best?} In terms of design features, the methods with the strongest performance in a label noise setting have a distance-based scoring function, and take features as input rather than class probabilities. SHE is the only OOD detector satisfying both criteria which doesn't sit at the top of the pile in Table~\ref{tab:best-ood-per-method} - we attribute its lower performance to two factors: it summarizes the ID dataset only with class-wise means (which may be overly reductive in a label noise setting where variance is larger), and it only considers correctly predicted samples when computing them (which may be small in number if the classifier is inaccurate or the number of classes is high). In contrast, GRAM which includes higher-order raw moments to describe ID data statistics is the top-1 method in Table~\ref{tab:best-ood-per-method}. In the comparison of Figure~\ref{fig:auroc_gap-vs-accuracy_gap_per_checkpoint}, GRAM and MDSEnsemble - the only methods in our benchmark which incorporate features at different depths in the network - stand out as having the ``flattest" accuracy-AUROC curves, which is especially beneficial when the training dataset is inherently difficult (e.g. CIFAR100 due to fine-grained labels or Clothing1M due to the image complexity and diversity). However, we note that the performance of MDSEnsemble and GRAM is highly architecture-dependent - the best OOD detection performance is achieved with a ResNet18 classifier, while MLPMixer and CCT architectures give sub-par results (often sub-50\% ie. even worse than a random detector). Whether this large performance variation is due to the layer types, feature dimensionality or other factors, and whether it can be remedied warrants further investigation.

\noindent \ul{\textbf{Takeaway: Distance is healthy}}
    Out of the 20 post-hoc OOD detectors in our benchmark, distance-based OOD detectors operating in feature space appear the most promising to cope with the problem of unreliable predictions. Intuitively, distance-based methods are more dissociated from the classifier's prediction, and more dependent on the content/appearance of ID images. In contrast, we did not find compelling evidence that methods targeting class logits or class probabilities for OOD detection are better suited for the noisy label setting.

\textbf{Are there tricks that work?} We consider 3 popular ``tricks'' aiming to better separate ID vs. OOD samples in logit or probability space - temperature scaling, input perturbation and sparsification - and assess their effectiveness in a noisy label setting (excluding cleanly trained models). To isolate the effect of Softmax temperature scaling and input perturbation, we introduce ODIN\textsubscript{notemp} (ODIN with temperature $T$ fixed to 1) and ODIN\textsubscript{nopert} (perturbation magnitude $m$ set to 0). We find that scaling $T$ by maximizing likelihood on ID validation labels is detrimental (\emin{MSP}{TempScale}{0.15}), however picking $T$ based OOD validation detection performance does make a statistically significant (though not practically significant) difference (\emin{ODIN\textsubscript{nopert}}{MSP}{0.05}). Input perturbation does not help in a label noise setting: looking at the optimal $m$ selected during ODIN\textsubscript{notemp}'s automatic parameter tuning, we observe that as label noise rate increases, the more likely that $m=0$ is picked (no perturbation). As for feature or weight sparsification, we note that REACT and DICE are the most promising logit-based methods in the AUROC\textsubscript{incorrect vs. OOD} ranking of Figure~\ref{fig:intro-fig}.

\subsection{Let's not forget about the validation set} \label{sec:val-stuff}

\textbf{Picking a model checkpoint} While it is well-understood that early stopping is beneficial to classification accuracy when training a classifier with noisy labels~\cite{early-stopping-label-noise-2020}, we investigate whether this extends to OOD detection performance.  We compare the OOD detection performance for the 2 checkpointing strategies, and find that for almost all methods, early stopping is beneficial (\emininf{early}{last}{0.5}). However, looking at Figure~\ref{fig:auroc_gap-vs-accuracy_gap_per_checkpoint}, we note that early stopping may increase the \textit{rate} at which OOD detection performance drops due to label noise for a given drop in accuracy - to an extreme in the case of TempScaling. A closer look at Figure~\ref{fig:noise-types-score-stats} gives some insight into its effect on the logits.

\textbf{What about OOD detector parameter tuning?} Many of the methods in our benchmark involve a set-up step where dataset-specific parameters are computed (e.g. statistics for ID samples) and/or a tuning step where hyperparameters are tuned to maximize OOD detection performance on a held-out validation OOD set. The set of (hyper)parameters for each method is outlined in the supplementary. Among these methods, some make use of \textit{classification labels} during set-up/tuning - e.g. to compute statistics for each class. In a label noise setting, this raises the question of whether to use a \textit{clean} validation set or the \textit{noisy} training set for set-up/tuning, or whether this makes a difference. We compare both settings for the 6 methods in our benchmark making use of class labels during set-up: MDS, RMDS, MDSEnsemble, GRAM, OpenMax and SHE, with results visualized in the supplementary. For SHE which computes the mean of features for each class during set-up, there is no statistically significant difference between using clean validation labels or potentially noisy training labels, although the latter may be better in some cases (\emin{SHE\textsubscript{val}}{SHE\textsubscript{train}}{1} and \emin{SHE\textsubscript{train}}{SHE\textsubscript{val}}{0.63}).  For methods based on the Malahanobis score, using noisy training labels to compute class-wise feature means and tied covariance is better (\emin{MDS\textsubscript{train}}{MDS\textsubscript{val}}{0.19} and \emin{RMDS\textsubscript{train}}{RMDS\textsubscript{val}}{0}) - intuitively, the class-specific statistics are more accurate with more data. Common to these 3 methods is that the OOD score at test-time does not depend on the \textit{predicted} class (likely to be incorrect in a label noise setting), but is rather based on distance to the \textit{closest} class in feature space (regardless of what class is predicted). OpenMax computes the mean logit per class, only considering correctly predicted samples (labels are used to check correctness) - using a potentially noisy training set yields consistently better performance (\emin{OpenMax\textsubscript{train}}{OpenMax\textsubscript{val}}{0}). Lastly, and in contrast to the other methods, GRAM benefits from using clean validation samples rather than a large number of noisy training samples for computing class-specific bounds of feature
correlations (\emin{GRAM\textsubscript{val}}{GRAM\textsubscript{train}}{0.23}). However, the performance gap between the two settings is small.

\noindent \ul{\textbf{Takeaway: Clean isn't always better or possible}}
  The use of clean vs. noisy labels during label-based parameter tuning is an important consideration. For distance-based methods which compute class-wise statistics, it appears that quantity often trumps quality, even when over 30\% of training labels are incorrect. This is promising for applications where a clean validation set is not available (e.g. medical imaging where labels are inherently subjective~\cite{dl-noisy-labels-med-2020}).
  
\subsection{What about a more realistic setting?}\label{sec:osr}

\begin{figure}[h]
    \centering
    \includegraphics[width=\columnwidth]{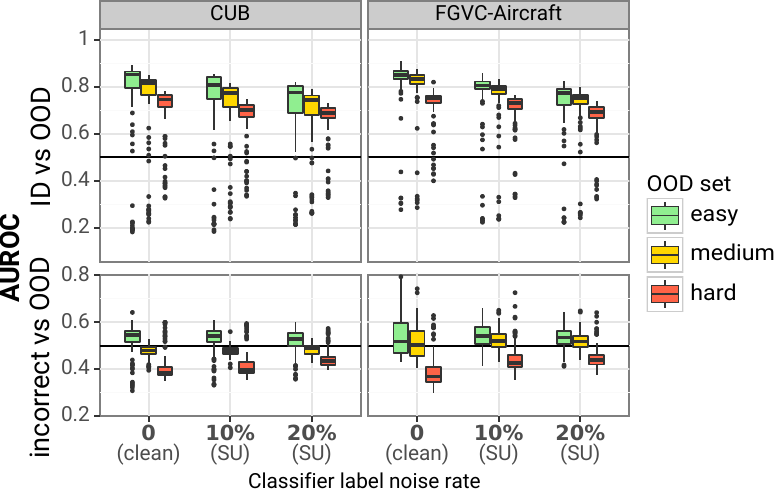}
    \caption{Each boxplot shows the performance distribution across 6 classifiers (3 seeds, 2 checkpoints) $\times$ 20 post-hoc methods, considering all ID test samples (top) or only incorrectly classified ones (bottom) in the AUROC metric.}
    \label{fig:osr-boxplot}
\end{figure}

We have thus far studied OOD detection in a simple (but standard~\cite{openood-benchmark-2022}) setting where the base classifier is trained from scratch, and where there is strong semantic and covariate shift between ID and OOD images. Yet in practice, pre-training is widely adopted, and distribution shifts may be much more subtle. We therefore extend our study of label noise to \textit{fine-grained semantic shift detection} with a base classifier that has been \textit{pre-trained on ImageNet}~\cite{imagenet-2009} before being trained on a dataset of interest. We follow the Semantic Shift Benchmark (SSB), where the goal is to detect \textit{unknown classes} from a known dataset (e.g. held-out bird species from the CUB~\cite{cub-dataset-2011} dataset or held-out aircraft model variants from FGVC-Aircraft~\cite{aircraft-dataset-2013}). Using SSB splits, we train ResNet50s (pre-trained) on half of the classes from CUB/FGVC-Aircraft (448x448 images), and we evaluate post-hoc OOD detection performance on known classes from the test set \textbf{(ID)} vs. the remaining unseen classes \textbf{(OOD)} split into 3 increasingly difficult sets. Since clean vs. real noisy label pairs are not available, we inject synthetic label noise in the training set (SU noise model) and follow the same evaluation procedure as in previous sections. Fig.~\ref{fig:osr-boxplot} summarizes its detrimental effect on fine-grained semantic shift detection across the 20 studied OOD detection methods: increasing label noise and ``difficulty'' of the OOD set act as orthogonal bottlenecks to detection performance. Increased label noise pulls AUROC\textsubscript{ID vs. OOD} and AUROC\textsubscript{incorrect vs. OOD} towards 50\%. 

\noindent\ul{\textbf{Takeaway: Limitations of post-hoc OOD detectors extend beyond toy settings}} Even in a more realistic setting where the base classifier has first been pre-trained on ImageNet and OOD samples are similar in appearance to the ID dataset, all 20 methods poorly separate incorrectly classifed ID samples from OOD samples, and degrade when the classifier has been trained on noisy labels. 

\section{Zooming out}

\textbf{Study limitations and possible extensions} We have focused on post-hoc OOD detection methods due to their pragmatic appeal and to maintain experimental feasibility. Extending this study to training-based OOD detection methods~\cite{openood-v1.5-2023} would of course be valuable. Aligning with OOD benchmarks~\cite{openood-benchmark-2022}, we also trained the base classifiers with a standard discriminative objective. Alternative supervision schemes may also be considered, and the effect of pre-training (and on what?) would be interesting to further analyse in a label noise setting, as it been shown to improve post-hoc OOD detection performance~\cite{pre-training-improves-robustness-2019,ood-effective-robustness-finetuning-2022,mim-ood-2023}. Lastly, the potential of noisy label removal~\cite{identifying-mislabeled-data-2020,filtering-noisy-instances-2021} or noise-robust learning~\cite{controlled-noisy-labels-2020,mitigation-noise-memorization-clipping-2023} techniques from the label noise literature (designed with classification performance in mind) for improving OOD detection would be a natural next step.

\textbf{Conclusion} We have explored the intersection between classification label noise and OOD detection, and conducted extensive experiments to extract new insights into the limitations of existing post-hoc methods. Our findings also echo the need to re-think the aims and evaluation of OOD detection in the context of safe deployment~\cite{reflect-eval-failure-detection-2023} (e.g. do we really want to exclude ID misclassifications from detection?). We hope that this work paves the way for future investigations which prioritize the robustness and applicability of OOD detection models in practical, imperfect classification scenarios which account for data uncertainty.

\section{Acknowledgements}

This work was supported by the Danish Data Science Academy, which is funded by the Novo Nordisk Foundation (NNF21SA0069429) and VILLUM FONDEN (40516). Thanks to the Pioneer Centre for AI (DNRF grant P1).

{
    \small
    \bibliographystyle{ieeenat_fullname}
    \bibliography{main}
}